\documentclass{article} 
\usepackage{iclr2021_conference,times}

\usepackage{amsmath,amsfonts,bm}

\def\eqref#1{equation~\ref{#1}}

\def\1{\bm{1}}

\DeclareMathAlphabet{\mathsfit}{\encodingdefault}{\sfdefault}{m}{sl}
\SetMathAlphabet{\mathsfit}{bold}{\encodingdefault}{\sfdefault}{bx}{n}

\usepackage[utf8]{inputenc} 
\usepackage[T1]{fontenc}    
\usepackage{hyperref}       
\usepackage{url}            
\usepackage{booktabs}       
\usepackage{nicefrac}       
\usepackage{microtype}      
\usepackage{graphicx}
\usepackage{amsmath}
\usepackage{xcolor}
\usepackage{mathtools}
\usepackage[ruled,linesnumbered,vlined]{algorithm2e}
\iclrfinalcopy

\newcommand{\modelshort}{RGM }
\newcommand{\algoname}{A2C + RGM }

\title{Beyond Exponentially Discounted Sum: Automatic Learning of Return Function}

\author{%
    Yufei Wang\thanks{equal contribution} \\
  Peking University\\
  \texttt{wang.yufei@pku.edu.cn} \\
   \And
   Qiwei Ye \footnotemark[1]\\
   Microsoft Research Asia \\
   \texttt{qiwye@microsoft.com} \\
   \And
    Tie-Yan Liu \\
   Microsoft Research Asia \\
   \texttt{tie-yan.liu@microsoft.com} \\
}

\begin{document}

\maketitle

\begin{abstract}
  In reinforcement learning, Return, which is the weighted accumulated future rewards, and Value, which is the expected return, serve as the objective that guides the learning of the policy. In classic RL, return is defined as the exponentially discounted sum of future rewards. One key insight is that there could be many feasible ways to define the form of the return function (and thus the value), from which the same optimal policy can be derived, yet these different forms might render dramatically different speeds of learning this policy. In this paper, we research how to modify the form of the return function to enhance the learning towards the optimal policy. We propose to use a general mathematical form for return function, and employ meta-learning to learn the optimal return function in an end-to-end manner. We test our methods on a specially designed maze environment and several Atari games, and our experimental results clearly indicate the advantages of automatically learning optimal return functions in reinforcement learning.

\end{abstract}


\section{Introduction}
The objective of reinforcement learning (RL) is to find a policy that takes the best action to gain accumulated long-term rewards, through trial and error. There are two important notions in RL: \emph{return}, which is the weighted accumulated future rewards after taking a series of actions, and \emph{value}, which is the expected value of the return, serveing as the objective for the policy function to maximize. In classic RL \citep{sutton1998introduction}, the return is usually defined as the exponentially discounted sum of future rewards, where the discounting factor $\gamma$ balances the importance of the immediate and future rewards. The motivation for using such a mathematical form of return is based on the economic theory~\citep{sutton1998introduction, pitis2019rethinking}, however, from the perspective of machine learning, there could be many different feasible ways to define the form of the return function (and thus the value), from which the same optimal policy can be derived. For example, let us consider the task of navigating through a maze. Any form of the return function that gives higher values to the correct path leading to the exit would generate the same optimal policy. However, these different forms might render dramatically different speeds of learning this policy. For example, suppose the agent can only receive a long-delayed non-zero reward when it reaches the exit. The return function that exponentially discounts this reward (inversely) along the path might incur very slow learning at the beginning of the correct path, while the return function that backs up this reward without decay might greatly ease the learning at the beginning. Based on this observation, our paper is concerned with the following question: Is there any better form of the return function than exponentially discounted sum, and can we design an algorithm to automatically learn such a function?

Actually, there have been some previous works that aim at finding better return/value functions that render more stable and faster policy learning. They can be mainly  classified into two categories. The first kind of works modify the form of the return function. For example, \cite{xu2018meta, zahavy2020self} proposed to use meta-learning to learn the hyperparameters {$\lambda$, $\gamma$} used in the TD-$\lambda$ methods. This generalizes the case of using fixed values of {$\lambda$, $\gamma$}, but in general the return function is still constrained in the exponentially discounted sum form. \cite{arjona2018rudder} employed a LSTM model to predict the delayed terminal reward at each time step, and then redistributed the delayed reward to previous time steps according to the prediction error (or the heuristic analysis on the hidden states of LSTM). However, there is no guarantee that the heuristic redistribution can lead to better performance or fast learning. A very recent work~\citep{xu2020meta} proposes to directly use a neural network to parameterise the update target, including value and returns, in reinforcement learning objectives. Our work differs from theirs in that we inject special structure and relational reasoning into the learned update target by employing a linear combination of rewards as the return and use a transformer~\citep{vaswani2017attention} to assign the linear coefficients based on the relationship of states in a trajectory.  The second kind of works focus on modifying the rewards used for return computation, and do not change the form of the return function at all. Examples include reward shaping \citep{ng1999policy}, reward clipping \citep{mnih2013playing}, intrinsic reward \citep{chentanez2005intrinsically,icarte2018using}, exploration bonus \citep{bellemare2016unifying}, and reward from auxiliary task \citep{jaderberg2016reinforcement}. In addition to these hand-crafted manipulations on rewards, a few recent works attempted to learn the manipulations automatically \citep{zheng2018learning,xu2018meta, veeriah2019discovery}.

In clear contrast to the aforementioned previous works, our work aims at automatically learning the appropriate mathematical form of the return function. Different from just learning the hyperparameters in the exponentially discounted sum, \textbf{we propose to use a general mathematical form for the return function, and employ meta learning to learn the optimal return function in an end-to-end manner.} In particular, we argue that the general form of the return function should take into account the information of the whole trajectory instead of merely the current state. This mimics the process of human reasoning, as we human always tend to analyze a sequence of moves as a whole. We implement our idea upon modern actor-critic algorithms \citep{babaeizadeh2017ga3c,schulman2017proximal}, and demonstrate its efficiency first in a specially designed Maze environment\footnote{https://github.com/MattChanTK/gym-maze}, and then in the high-dimensional Atari games using the Arcade Learning Environment(ALE) \citep{bellemare13arcade,machado17arcade}. The experimental results clearly indicate the advantages of automatically learning optimal return functions in reinforcement learning.

The paper is organized as follows. In Section \ref{sec::preliminaries} we introduce related works and background settings. In Section \ref{sec::methods}, we formalize the proposed idea and describe the meta learning algorithm in detail. In Section \ref{sec::exp1}, the experimental results are presented. The paper concludes in Section \ref{sec::conclusion}.


\section{Background and Related Work}\label{sec::preliminaries}

\subsection{Modifying The Mathematical Form of Return Function}
Lots of works has researched the discounting factor in the exponentially discounted form return function. \cite{lattimore2011time} proposed to generalize the exponentially discounting form through discount functions that change with the agents' age. Precisely choosing different $\gamma$ at different time has been proposed in \citep{franccois2015discount,OpenAI_dota}. \cite{xu2018meta} treated hyper-parameter [$\lambda$, $\gamma$] in TD-$\lambda$ as learnable parameters that could be optimized by meta-gradient approach. \cite{pitis2019rethinking} introduced a flexible state-action discounting factor by characterizing rationality in sequential decision making. \cite{barnard1993temporal} generalized the work of \cite{singh1992scaling} to propose a multi-time scale TD learning model, based on which \cite{romoff2019separating,reinke2017average,sherstan2018generalizing} explored to decompose the original discounting factor into multiple ones. \cite{arjona2018rudder} proposes to decompose the delayed return to previous time steps by analyzing the hidden states of a LSTM.
All these works suggest that the form of return function is one of the key elements in RL. Our work generalizes them by proposing to use a general form return function.

\subsection{Reward Augmentation}
Another family of works focus on manipulating the reward functions to enhance the learning of the agents.  
Reward shaping \citep{ng1999policy} shows what property a reward modification should possess to remain the optimal policy, i.e., the potential-based reward. Lots of works focus on designing intrinsic reward to help learning, e.g., \citep{oudeyer2009intrinsic,schmidhuber2010formal,stadie2015incentivizing} used hand-engineered intrinsic reward that greatly accelerate the learning. Exploration bonus is another class of works that add bonus reward to improve the exploration behaviours of the agent so as to help learning, e.g., \cite{bellemare2016unifying,martin2017count,tang2017exploration} propose to use the pseudo-count of states to derive the bonus reward. \citep{sutton2011horde,mirowski2016learning,jaderberg2016reinforcement} propose to augment the reward by considering information coming from the task itself. 

\subsection{Learning to Learn in Reinforcement Learning}
Metagradient~\citep{zahavy2020self, xu2018meta, zheng2018learning, xu2020meta, oh2020discovering} is a general method for adapting some of
the hyperparameters in the learning algorithm online. 
This approach has
been used to tune many different meta-parameters of RL algorithms in various settings, such as the $\gamma$ and $\lambda$ in TD-$\lambda$~\citep{xu2018meta}, intrinsic rewards~\citep{zheng2018learning}, auxiliary tasks~\citep{veeriah2019discovery}, and etc. For a summary of different learning to learn methods in RL, we refer the readers to~\citep{xu2020meta}. We also use metagradients in our paper, to learn the linear coefficients in a new form of return function.


\section{Methods}\label{sec::methods}

\subsection{Classic RL Settings}
Here we briefly review the classic RL settings, where return and value are defined as the exponentially discounted sum of accumulated future rewards. We will also review the policy gradient theorem, upon which our method is built on. We assume an episodic setting. At each time step $t$, the agent receive a state $s_t $ from the state space $S$, takes an action $a_t$ in the action space $A$, receives a reward $r_t$ from the environment based on the reward function $r_t = r(s_t, a_t)$, and then transfers to the next state $s_{t+1}$ according to the transition dynamics $P: S \times A \times S \rightarrow [0,1]$. Given a trajectory $\tau = (s_0, a_0, r_0, ..., s_T, a_T, r_T, ...)$, the \textbf{return} of $\tau$ is computed in the following exponentially discounted sum form: $G(\tau) = \sum_{t=0}^{\infty}\gamma^{t}r_{t}$, , where $\gamma$ is the discounting factor. Similarly, the return of a state-action pair $(s_t, a_t)$ is computed as: $G(s_t, a_t) = \sum_{l=t}^{\infty}\gamma^{l-t}r_{l}$.

A policy $\pi$ (which is usually parameterized by $\theta$, e.g., a neural network) is a probability distribution on $A$ given a state $s$, $\pi_{\theta}: S \times A \rightarrow [0, 1]$. A trajectory $\tau$ is generated under policy $\pi_{\theta}$ if all the actions along the trajectory is chosen following $\pi_{\theta}$, i.e., $\tau \sim \pi_{\theta}$ means $a_t \sim \pi_{\theta}(\cdot | s_t)$ and $s_{t+1} \sim P(\cdot | s_t, a_t)$.   
Given a policy $\pi_{\theta}$, the \textbf{value} of a state $s$ is defined as the expected return of all trajectories when the agent starts at $s$ and then follows $\pi_\theta$: $V^{\pi_{\theta}}(s) = E_{\tau}[G(\tau) | \tau(s_0) = s, \tau \sim \pi_\theta]$. Similarly, the value of a state-action pair is defined as: $Q^{\pi_{\theta}}(s,a) = E_{\tau}[G(\tau) | \tau(s_0) = s, \tau(a_0) = a, \tau \sim \pi_\theta]$. 

The performance of a policy $\pi_{\theta}$ can be measured as $J(\pi_{\theta}) = E_{\tau}[G(\tau) | \tau(s_0) \sim \rho, \tau \sim \pi_\theta]$, where $\rho$ is the initial state distribution. For brevity, we will write $J(\theta) = J(\pi_{\theta})$ throughout. The famous policy gradient theorem \citep{sutton2000policy} states:
\begin{equation}\label{eq::pgt}
    \frac{dJ(\theta)}{d\theta} = E_{s_t \sim P(\cdot | s_{t-1}, a_{t-1}), a_t \sim \pi_{\theta}(\cdot | s_t)}[\nabla_{\theta} \text{log}\pi_{\theta}(a_t | s_t) Q^{\pi_{\theta}}(s_t, a_t)]
\end{equation}
Usually, a baseline value $B(s_t)$ can be subtracted by $Q^{\pi_{\theta}}(s_t, a_t)$ to lower the variance, e.g., in A3C \citep{mnih2016asynchronous} the state value $V^{\pi_\theta}(s_t)$ is used.
In the following, for the simplicity of derivation, we will assume that $Q^{\pi_{\theta}}(s_t, a_t)$ is computed using a sample return $G(s_t, a_t) = \sum_{l=t}^{\infty}\gamma^{l-t}r_{l}$.

\subsection{Learning a General Form of Return Function}
\label{subsec:new return}
The key of our method is that we do not restrict the function form of return to be just the exponentially discounted sum of future rewards. Instead, we compute it using an arbitrary function $g$, which is parameterized by a neural network $\eta$, and we call $g_\eta$ as the \textbf{return generating model}. In addition, the function $g$ considers the information of the whole trajectory: 
\begin{equation}
    G^g(s_t, a_t) \vcentcolon = g_{\eta}(t, s_0, a_0, r_0, ..., s_T, a_T, r_T, ...)
    \label{eq::newreturn}
\end{equation}
 We can then optimize our policy $\theta$ with regard to this new return by substituting it in the policy gradient theorem, where $\alpha$ is the learning rate:
\begin{equation}
\begin{split}
    \hat{\frac{dJ(\theta)}{d\theta}} &= E_{s_t, a_t \sim \pi_{\theta}}[\nabla_{\theta} 
    \text{log}\pi_{\theta}(a_t | s_t) G^{g}(s_t, a_t)],~~~~~~
    \theta' = \theta + \alpha \hat{\frac{dJ(\theta)}{d\theta}}
\end{split}
\label{eq::newpg}
\end{equation}

We now show how to train $g_\eta$, so that it renders faster learning when the policy is updated using the new return. 
For the simplicity of derivation, we take $g$ in the following particular form:
\begin{equation}
\begin{split}
     G^{g}(s_t, a_t) &= \sum_{l = t}^{\infty} \beta_l^\eta r_{l},~~~~~~~~
     \beta_l^\eta = g_\eta(l, s_0, a_0, r_0, ..., s_T, a_T, r_T, ...), ~~ t = 1, ..., T, ...
\end{split}
\label{eq::linearcombine}
\end{equation}
i.e., the new return is calculated as a linear combination of future rewards, and the linear coefficient is computed using the return generating model $g_\eta$. 

Using meta-learning, it is straightforward to set the learning objective of  $g_\eta$ to be the performance of the updated policy $\pi_\theta'$, and thus $g_\eta$ can be trained using chain rule. The similar idea has also been used in \cite{xu2018meta} and \cite{zheng2018learning}. Formally, with the particular linear combination form of the new return, update in E.q.\ref{eq::newpg} becomes:
\begin{equation}
\begin{split}
    G^{g}(s_t, a_t) &= \sum_{l = t}^{\infty} \beta_l^\eta r_{l},~~~~~~~~~~~~~ 
    \theta' = \theta + \alpha \nabla_{\theta} \text{log}\pi_{\theta}(a_t | s_t) G^g(s_t, a_t)
\end{split}
\label{eq::updatewithnewretlinear}
\end{equation}
We want the new return $G^{g}$ to maximize the effect of such an update, i.e., to maximize the performance of the updated policy $\pi_\theta'$. By the chain rule, we have:
\begin{equation}
    \frac{dJ(\theta')}{d\eta} = \frac{dJ(\theta')}{d\theta'} \frac{d\theta'}{d\eta}
    \label{eq::chainrule}
\end{equation}
The first term $\frac{dJ(\theta')}{d\theta'}$ in E.q. \ref{eq::chainrule} can be computed by policy gradient theorem (E.q \ref{eq::pgt}), with the \textit{original return and samples collected under the new policy $\pi_\theta'$}:
\begin{equation}
\frac{dJ(\theta')}{d\theta'} = E_{s', a' \sim \pi_{\theta'}}[{\nabla_{\theta'} \text{log}\pi_{\theta'}(a'| s') G(s', a')}]
\label{eq::newthetapg}
\end{equation}
and the second term $\frac{d\theta'}{d\eta}$ in E.q. \ref{eq::chainrule} can be computed using E.q. \ref{eq::updatewithnewretlinear}:
\begin{equation}
    \begin{split}
        \frac{d\theta'}{d\eta} &= \alpha \nabla_{\theta} \text{log}\pi_{\theta}(a_t | s_t) \frac{dG^g(s_t, a_t)}{d\eta} = \alpha \nabla_{\theta} \text{log}\pi_{\theta}(a_t | s_t) \sum_{l=t}^{\infty} r^l \frac{d\beta_l^\eta}{d\eta}
    \end{split}
\label{eq::dtheta'deta}
\end{equation}
Therefore, the gradient $\frac{dJ(\theta')}{d\eta}$ can be computed by combining E.q. \ref{eq::newthetapg} and E.q. \ref{eq::dtheta'deta}, and any first-order methods can be used to optimize $g_\eta$ (e.g., gradient ascent).

\begin{figure}[h]
\begin{minipage}{.5\linewidth}
\begin{algorithm}[H]
    \small
	\SetKwProg{Fn}{Function}{:}{}
    \SetKwData{Input}{Input}{}
    \SetKwData{Output}{Output}{}
    \DontPrintSemicolon
    \Input : policy network $\pi_\theta$, reward generating model $g_\eta$, training iterations $N$, step size $\alpha$ and $\alpha'$ \\
    \For{$i = 1$ \KwTo $N$}
    {
      Sample a trajectory $\tau = \{s_0, a_0, r_0, ..., s_T, a_T, r_T\}$ with length $T$ using $\pi_\theta$  \\
      Compute the new return $G^{g}(s_t, a_t)$ using $g_\eta$ as in E.q. \ref{eq::linearcombine} \\
      update $\theta' = \theta + \alpha \nabla_{\theta} \text{log}\pi_{\theta}(a_t | s_t) G^\eta(s_t, a_t)$ as in E.q. \ref{eq::updatewithnewretlinear} \\
      Approximate $\frac{dJ(\theta')}{d\theta'}$ using E.q. \ref{eq::importancesamplednextpg} \\ 
      Compute $\frac{d\theta'}{d\eta}$ using E.q. \ref{eq::dtheta'deta} \\
      Compute $\frac{dJ(\theta')}{d\eta} = \frac{dJ(\theta')}{d\theta'}\frac{d\theta'}{d\eta}$ \\
      update $\eta' \leftarrow \eta + \alpha' \frac{dJ(\theta')}{d\eta}$
    }	
    \caption{Learning Return Computation with Meta Learning}
    \label{algo::lbr}
\end{algorithm}
\hspace{5cm}
\end{minipage}
    \begin{minipage}{.5\linewidth}
    \centering
    \includegraphics[width = .7\textwidth]{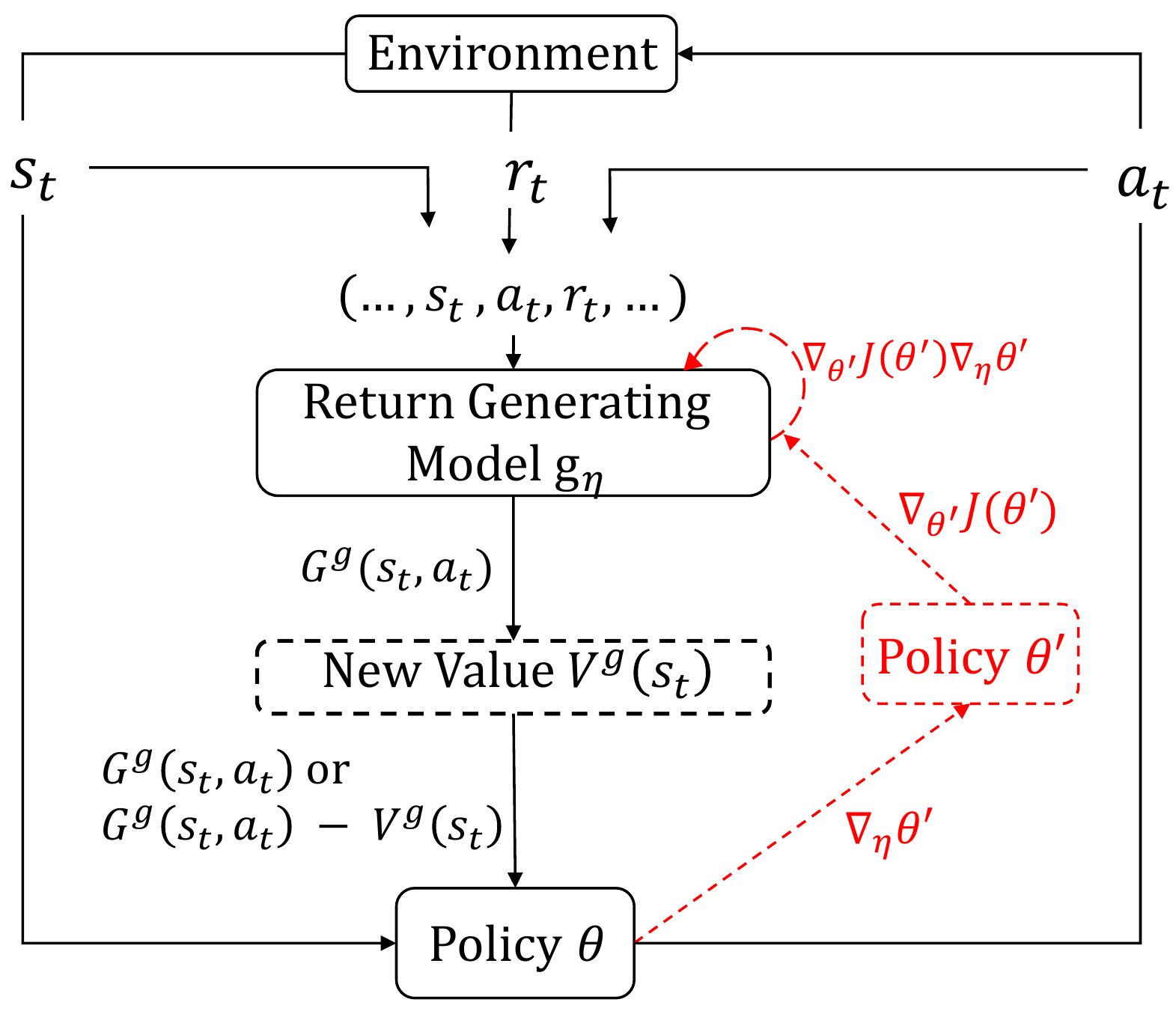}
    \caption{Relationship between Return Generating Model and existing policy gradient-based RL. We can optionally use a new value network to learn the new return generated by $g_\eta$, as shown by the dashed black box. The dashed red lines and box show how the return generating model $g_\eta$ can be trained using meta-learning and the chain rule.}
    \label{fig::rrmmodel}
    \end{minipage}
\end{figure}

In practice, when computing $\frac{dJ(\theta')}{d\theta'}$, we do not really sample a new trajectory using the updated policy as this increases the sample complexity. Instead, we compute it using the current trajectory with importance sampling:
\begin{equation}
    \frac{dJ(\theta')}{d\theta'} =  G(s_t, a_t) \nabla_{\theta'}\frac{ \pi_{\theta'}(a_t | s_t)}{\pi_\theta(a_t | s_t)}
    \label{eq::importancesamplednextpg}
\end{equation}
 The full procedure of our method is shown in Algorithm \ref{algo::lbr}, and a overview of the relationship between our return generating model and existing policy gradient-based RL algorithms is shown in figure \ref{fig::rrmmodel}.

For the clearance of illustration we only show our algorithm in the simplest case, i.e, 1) using the linear combination of future rewards as the new form of return function; 2) using vanilla gradient ascent to update $\theta$, and 3) using sample return $G(s, a)$ to approximate the Q-value $Q^{\theta}(s,a)$. 
Actually, our method can be applied much more broadly: 1) $g$ can take in any form as long as it is differentiable to the parameter $\eta$; 2) the update of the policy $\theta$ can employ any modern optimizer like RMSProp \citep{tieleman2012lecture} or Adam \citep{kingma2014adam}, as long as the update procedure of computing $\theta'$ is fully differentiable to $\eta$ (as in E.q. \ref{eq::dtheta'deta}); and 3) our methods can be combined with any advanced actor-critic algorithms like A3C \citep{mnih2016asynchronous} or PPO \citep{schulman2017proximal}, as long as the new gradient $\frac{dJ(\theta')}{d\theta'}$ can be effectively evaluated (as in E.q. \ref{eq::newthetapg}).

\subsection{Discussion}
\label{subsec::method-discussion}
\textbf{Viewing in Trajectory}. In this subsection we present how we design the structure of the return generating model $g_\eta$. Our first design principle is that $g_\eta$ should be able to use the information of the whole trajectory $(s_0, a_0, r_0, ..., s_T, a_T, r_T)$ to generate return. This mimics the process of human reasoning, as we always tends to analyze a sequence of moves. Our second design principle is that the model should be excel to analyze the relationship between different time steps when generating the return, which is again natural to human reasoning. With these two design principles, we choose to employ the Multi-Head Attention Module~\citep{vaswani2017attention} as a main block of our model. We adopted the encoder part of the Transformer architecture \cite{vaswani2017attention}.  Specifically, the input to our model is a whole trajectory $\tau = (s_0, a_0, r_0, ..., s_T, a_T, r_T)$. The model takes the trajectory $\tau$ as input, embeds it into a vector $e$ of dimension $d$, and then passes it through several stack of layers that consist of multi-head attention and residual feed forward connection. At last, the model uses a feed forward generator to generate the new returns $G^g(s_0, a_0), ..., G^g(s_T, a_T)$. In the linear combination case, the generator uses softmax to produce the normalized linear coefficient $\beta_1^\eta, ..., \beta_t^\eta$.

\begin{figure*}[h]
    \centering
    \includegraphics[width = 0.7\textwidth]{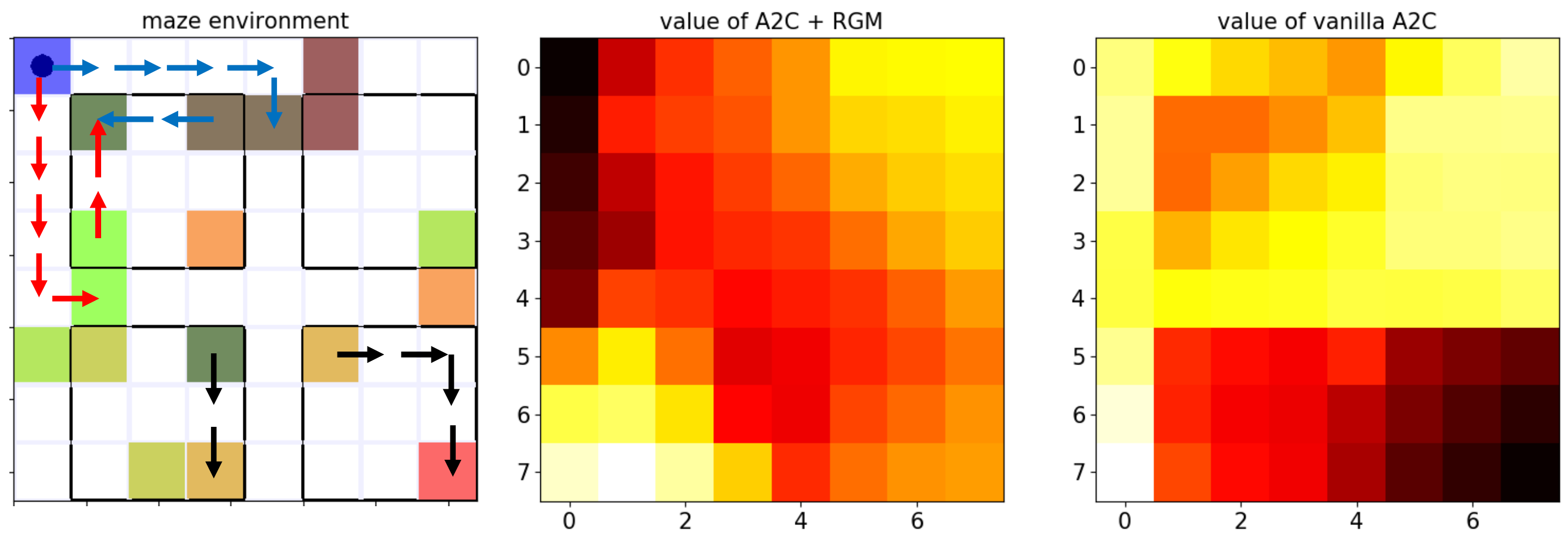}
    \caption{The values of grids of vanilla A2C (the value learns towards the traditional discounted sum return) and our \algoname (the value learns towards to the return generated by \modelshort). Darker color represents larger value. The value of A2C + RGM derives the correct path labeled by the red + black arrows in the leftmost panel, and the value of vanilla A2C derives another correct path labeled by blue + black arrows in the leftmost panel. }
    \label{fig:mazevalue}
\end{figure*}

\begin{figure*}[h]
\centering
\begin{tabular}{cc}
\includegraphics[width = 0.36\textwidth]{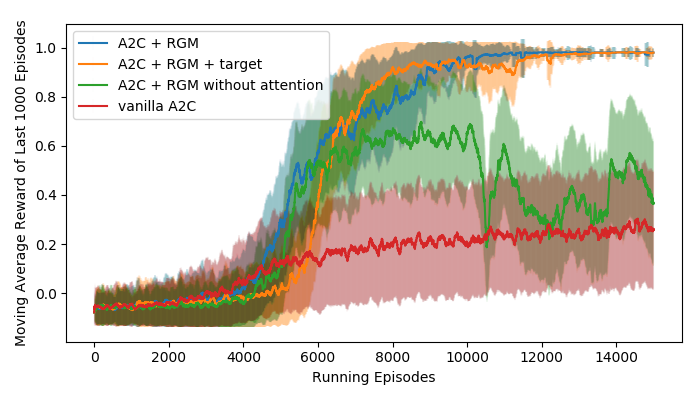}  &
    \includegraphics[width = 0.4\textwidth]{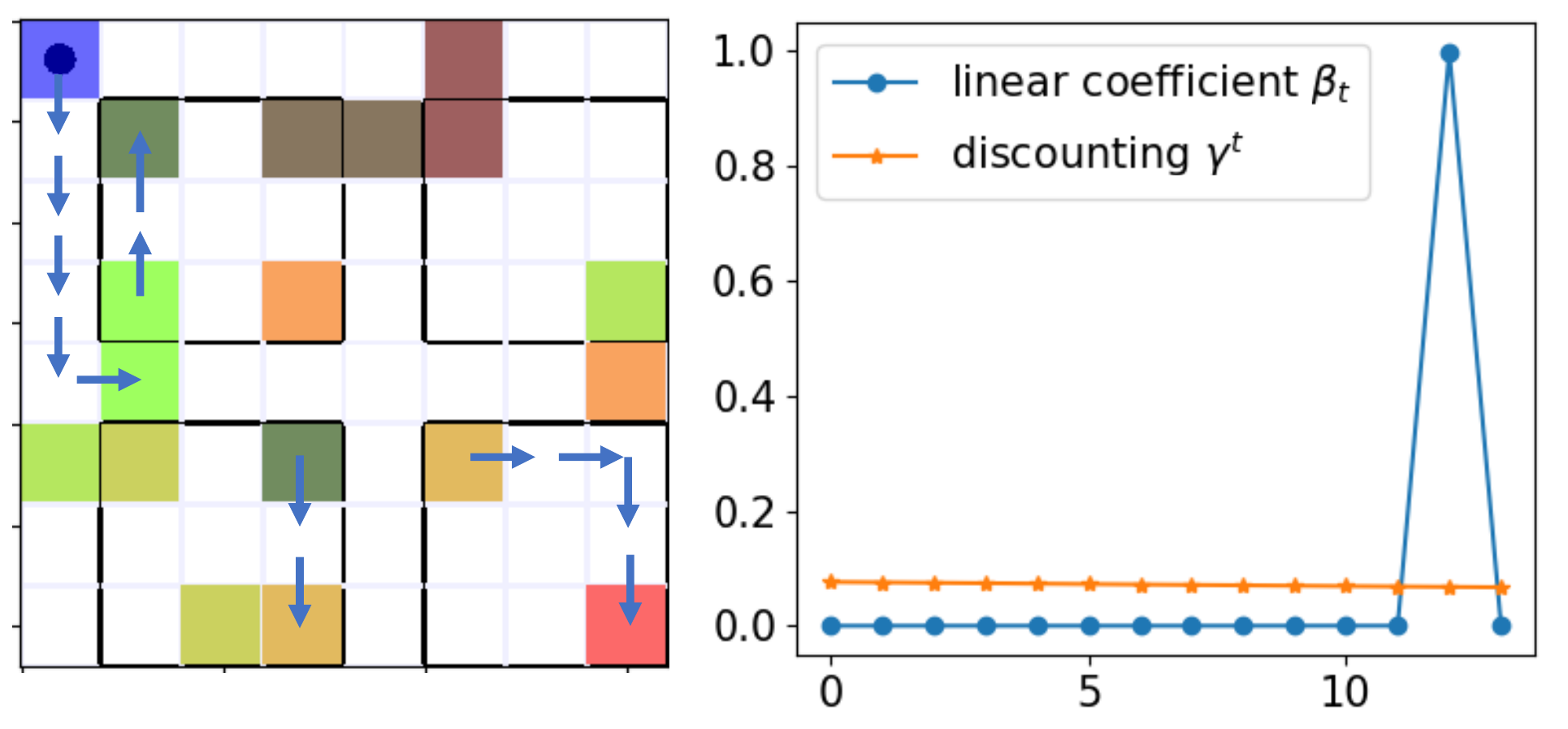}
\end{tabular}
    \caption{Left: Learning curves of the compared algorithms. The Y-axis is the moving average reward of the last 1000 episodes, and the shaded region is half a standard deviation. Middle and Right: Visualization of $\gamma^t$ and $\beta_t$ along the correct path, which is labeled by the arrows in the left maze.}
    \label{fig:trajvisual&learning curve}
\end{figure*}

\begin{figure*}[ht]
	\centering
	\begin{tabular}{cc}
	    \includegraphics[width=0.4\textwidth]{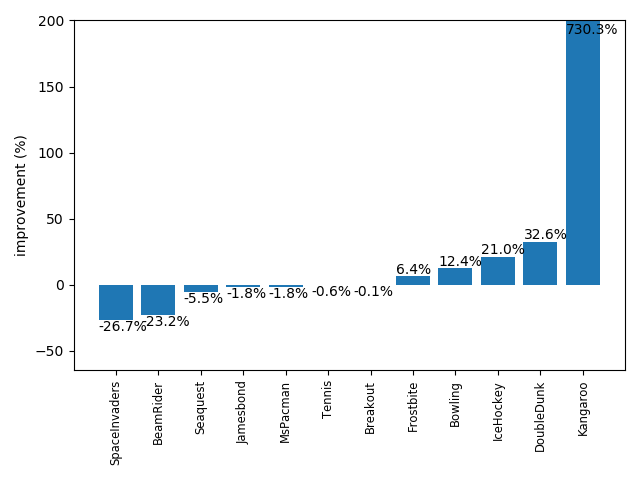}  &  \includegraphics[width=0.4\textwidth]{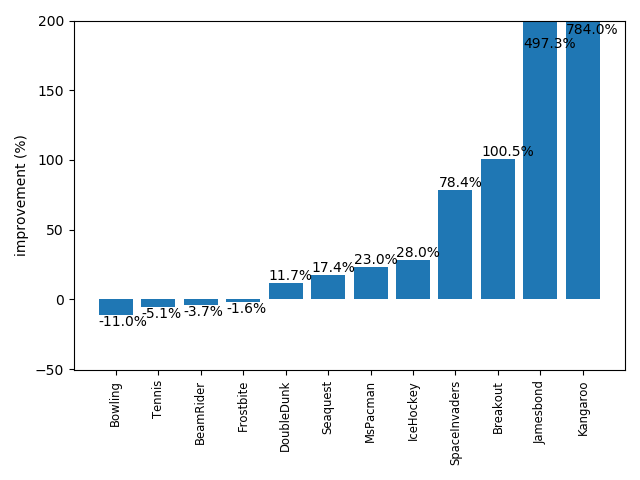} \\
	\end{tabular}
   
	\caption{Left: Improvements of A2C + RGM over vanilla A2C. Right: Improvements of A2C + RGM over LIRPG. The y-axes show relative raw game score improvements.}
    \label{fig:improvement}
\end{figure*}

\begin{figure*}[ht]
	\centering
    \includegraphics[width=.9\textwidth]{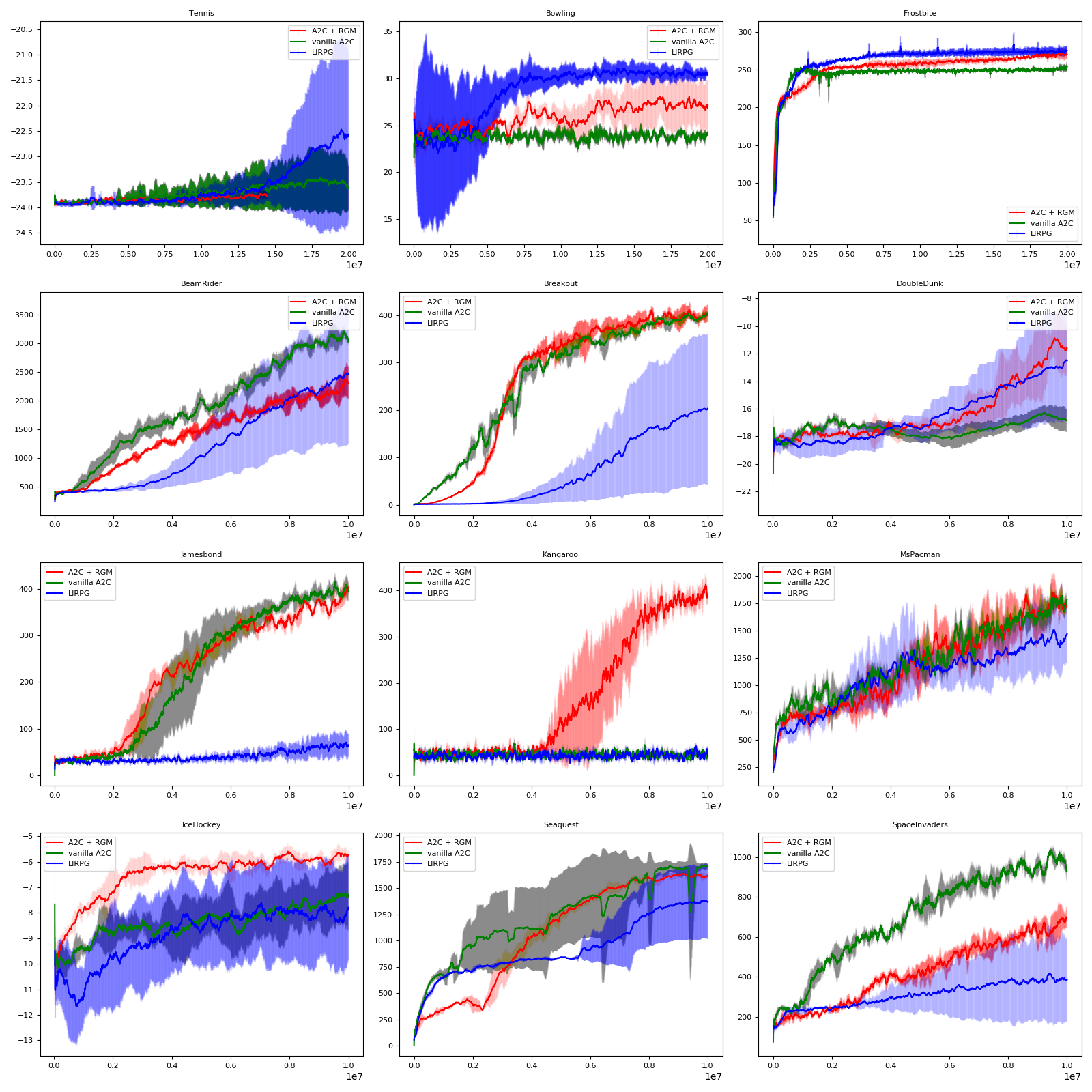}
	\caption{Comparison of Vanilla A2C and A2C + \modelshort on 12 Atari games. Red: A2C + RGM; Green: vanilla A2C. Blue: LIRPG. The X-axis is the trained time steps. The Y-axis is the raw game scores. The solid line is the mean of the raw scores in the last 100 episodes, averaged over 3 random seeds, and the shaded area is a standard deviation. As the large delayed-reward games Bowling, Tennis and Frostbite are very difficult to learn, we trained all algorithms on them for more time steps.}
    \label{fig:atari}
\end{figure*}

\textbf{From New Return to New Value.} The new return function naturally brings about a brand new value function. Indeed, we can similarly define the value of $V^g(s)$ and $Q^g(s,a)$ as the expected value of the new returns of trajectories starting from $s$ (taking action $a$), and following policy $\pi_\theta$:
\begin{equation}
    \begin{split}
        G^g(\tau) &= G^g(s_0, a_0) = g_\eta(0, s_0, a_0, r_0, ..., s_T, a_T, r_T, ...) \\
        V^g(s) = E_{\tau \sim \pi_\theta}[G^g(\tau) &| \tau(s_0) = s],~~~~~~~~~~~
        Q^g(s, a) = E_{\tau \sim \pi_\theta}[G^g(\tau) | \tau(s_0) = s, \tau(a_0) = a] \\
    \end{split}
\end{equation}
Here we give a very shallow analysis on this new value definition (one obvious future work would be to give a deep analysis on such definitions). The new value definition is way more general than the original one, but this generality also loses useful properties, e.g., the Bellman Equation might no longer applies. In the classic RL setting, the bellman equation tells that the value of the current state can be computed by bootstrapping from the value of the next state $V(s_t) = r_t + \gamma E_{s_{t+1}}[V(s_{t+1})]$, and the correction of such bootstrap might not hold true for an arbitrary return function $g$. As a result, temporal difference learning methods (e.g. Q-learning, SARSA) might not be applicable with the new value
On the other hand, the Monte-Carlo methods for evaluating the new value would still work, which only relies on the Law of Big Number.



\section{Experiments}\label{sec::exp1}

\subsection{General Setups}
In this subsection, we describe some general setups for the following experiments. We implemented our \modelshort upon the A2C \citep{mnih2016asynchronous} algorithm. All the experiment are running on a machine equipped with Intel(R) Xeon(R) CPU E5-2690, and four Nvidia Tesla M40 GPUs. In the following, we will denote the A2C baseline as the vanilla A2C, and denote our algorithms as A2C + \modelshort (short for Return Genearting Model). In both the illustrative case and the Atari experiment, the new form of return function is the linear combinations of the future rewards, as discussed in Section \ref{subsec:new return} and E.q. \ref{eq::linearcombine}. The \modelshort $g_\eta$ has 4 stacking layers and 4 heads in multi-head attention.  

\subsection{Illustrative Case}
In this section, we show our algorithm using an illustrative maze example. As shown in the leftmost panel of figure \ref{fig:mazevalue}, the maze~\citep{mattchantk_2016} is a 2-D grid world of size $8 \times 8$. The agent starts from the left-top corner, and needs to find its way to the exit at the bottom-right corner. Black lines in the maze represents unbreakable walls. The key feature of the maze is that it has \textit{portals}, shown as colored grids in figure \ref{fig:mazevalue}, which can transport the agent immediately from one location to another with the same color. Indeed, we deliberately created four isolated rooms in the maze, and to successfully reach the exit requires the agent to utilize these portals to transport between different rooms. Two possible routes have been marked out in figure \ref{fig:mazevalue}.

At each time step, the agent gets a state which is the coordinate of his current location, chooses an moving direction among up, down, left and right, and receives a reward of $-\epsilon$  if he does not reach the exit, and $1$ if he reaches the exit, which also ends the game. We set $\epsilon$ to be very small, e.g., $-0.01$ in our experiments. Therefore, the maze renders an environment that has very delayed reward. Besides, from our human beings' perspective, we would pay special attention to the portals, as they enable nonconsecutive spatial location change. 

We compare the vanilla A2C algorithm and our A2C + \modelshort algorithm. We use separate value and policy networks, both of them are feed-forward MLPs with three hidden layers of 64 neurons and the ReLU nonlinear activation. Figure \ref{fig:trajvisual&learning curve} (b) compares the learning curves of these two algorithms. For ablation study, we implemented two more versions of our \algoname besides the vanilla one. 1) `\modelshort + target': since our \modelshort is always learning (changing) itself, the learning objective it provides to the value network is also changing, which might cause oscillation for the learning of the value network. To alleviate this, similar to that in DQN \citep{mnih2013playing} and DDPG \citep{lillicrap2015continuous}, we use a target network of the \modelshort model and optimize the value network towards the target network to stabilize its learning. The parameters of the target network is copied from the learning \modelshort periodically.  2) `\modelshort without attention': we replaced all the attention modules with feed-forward layers to test whether the attention module is necessary. The results in figure \ref{fig:trajvisual&learning curve} (b) shows that our algorithm learns dramatically faster and better than the vanilla A2C algorithms. It also shows that the attention module is necessary for stable learning, and whether using a target network does not cause huge difference in learning in the maze environment.

To have a deeper understanding of how the \modelshort works, we visualized the values of the grids under the vanilla A2C's value network and our A2C + RGM's value network. The result is shown in figure \ref{fig:mazevalue}. As expected, the values of grids learned in vanilla A2C exponentially decreases as the distance to the exit increases, and there is no difference in values between the transport grids and vanilla grids. However, the values learned by our \modelshort is quite different. It shows that a heavy part of return is redistributed to the beginning of the correct path that leads to the exit, and the value almost follows a decreasing trend as it approaches the exit. Also, compared with the vanilla values, the portals in our \algoname tends to have higher values than the vanilla grids around it. Both values give the correct policy, as they all have higher values along the correct path. However, as indicated by Figure \ref{fig:trajvisual&learning curve} (b), these two different values render large difference in the learning speed towards the optimal policy.

To further investigate how the new value is learned, we visualized the linear coefficient $\beta_t$ generated by our \modelshort along the correct path, and compared it with the traditional discounted $\gamma^t$. The result is shown in figure \ref{fig:trajvisual&learning curve} (b). We normalize both coefficients to make them sum up to $1$. With $\gamma = 0.99$, the discounted form return gives almost equal favour to each reward encountered along the path. On the contrast, our \modelshort distributed almost all the weight to the delayed reward at the last time step, which is the only positive reward the agent receives when reaching the exit. As a result, when using the traditional discounted return, the delayed reward (in our setting, which is also the only effective learning signal) is exponentially decayed $13$ times before it can reach the initial state, which greatly hurts its propagation; on the other hand, our \modelshort could propagate back this learning signal with nearly zero loss to all previous states (recall that our \modelshort compute return as $G^g(s_t, a_t) = \sum_{t=0}^T \beta_t r_t$, if some $\beta_t$ is near 1, then $r_t$ will be fully used when computing all $G(s_i, a_i), ~~i \leq t$). With this effective back propagation of the learning signal, our \modelshort greatly eases the learning of the agents.

\subsection{Deep Reinforcement Learning Results on Atari Games}\label{sec::exp2}

We also tested our \algoname on multiple Atari games from the Arcade Learning Environment (ALE) \cite{bellemare2013arcade}. We implemented the baseline A2C algorithm using Pytorch with exactly the same network architecture as in \cite{mnih2016asynchronous}, and trained it using the same hyper-parameters as in the OpenAI implementation \cite{dhariwal2017openai}.
We do not use a separate target network for the RGM as it does not bring significant help in the maze experiments.

We compare our \algoname, with the vanilla A2C and LIRPG \cite{zheng2018learning}, in which the authors used the meta-learning methods to learn an intrinsic rewards at each time step instead of augmenting the return function.
Figure \ref{fig:improvement} shows the improvements of A2C + RGM over the baseline methods, and
Figure \ref{fig:atari} shows the learning curves of all games.
Among them, Bowling, Tennis, and Frosbite have very rare and delayed rewards, e.g., in Frostbite it takes the agent 180 time steps to get its first reward. DoubleDunk, IceHockey, Jamesbond and Kangaroo has an itermediate level of delayed rewards, e.g., the agent needs roughly 50 steps to get its first reward. The rest of the games, i.e., Seaquest, BeamRider, SpaceInvaders, Breakout and Mspacman all have rich immediate rewards. Using the original discounted sum return, learning in the delayed reward games are much more difficult than the rich-reward games. As shown in Figure \ref{fig:atari}, as expected, our \algoname brings large benefits in the delayed games by learning a better form of return computation. For the non-delayed reward games, it  boosts the  performance in some of them, while hurts the performance in some others, as the original exponentially discounted return is already good enough for learning in such non-delayed games. 
For the RGM, there are two hyper-parameters: the learning rate $\alpha'$ and the trajectory length $T$. We searched the following combinations of them
$\{1e-4, 1e-5, 1e-6, 1e-7\} \times \{50, 100, 200\}$, and plotted the best results from the search.

\section{Conclusion}\label{sec::conclusion}
Return and value serve as the key objective that guide the learning of the policy. One key insight is that there could be many different ways to define the computation form of the return (and thus the value), from which the same optimal policy can be derived. However, these different forms could render dramatic difference in the learning speed. In this paper, we propose to use arbitrary general form for return computation, and designed an end-to-end algorithm to learn such general form to enhance policy learning by meta-gradient methods. We test our methods on a specially designed maze environment and several Atari games, and the experimental results show that our methods effectively learned new return computation forms that greatly improved learning performance.

\bibliography{00_bibliography}

\begin{thebibliography}{43}
\providecommand{\natexlab}[1]{#1}
\providecommand{\url}[1]{\texttt{#1}}
\expandafter\ifx\csname urlstyle\endcsname\relax
  \providecommand{\doi}[1]{doi: #1}\else
  \providecommand{\doi}{doi: \begingroup \urlstyle{rm}\Url}\fi

\bibitem[Arjona-Medina et~al.(2018)Arjona-Medina, Gillhofer, Widrich,
  Unterthiner, and Hochreiter]{arjona2018rudder}
Jose~A Arjona-Medina, Michael Gillhofer, Michael Widrich, Thomas Unterthiner,
  and Sepp Hochreiter.
\newblock Rudder: Return decomposition for delayed rewards.
\newblock \emph{arXiv preprint arXiv:1806.07857}, 2018.

\bibitem[Babaeizadeh et~al.(2017)Babaeizadeh, Frosio, Tyree, Clemons, and
  Kautz]{babaeizadeh2017ga3c}
Mohammad Babaeizadeh, Iuri Frosio, Stephen Tyree, Jason Clemons, and Jan Kautz.
\newblock Reinforcement learning thorugh asynchronous advantage actor-critic on
  a gpu.
\newblock In \emph{ICLR}, 2017.

\bibitem[Barnard(1993)]{barnard1993temporal}
Etienne Barnard.
\newblock Temporal-difference methods and markov models.
\newblock \emph{IEEE Transactions on Systems, Man, and Cybernetics},
  23\penalty0 (2):\penalty0 357--365, 1993.

\bibitem[{Bellemare} et~al.(2013){Bellemare}, {Naddaf}, {Veness}, and
  {Bowling}]{bellemare13arcade}
M.~G. {Bellemare}, Y.~{Naddaf}, J.~{Veness}, and M.~{Bowling}.
\newblock The arcade learning environment: An evaluation platform for general
  agents.
\newblock \emph{Journal of Artificial Intelligence Research}, 47:\penalty0
  253--279, jun 2013.

\bibitem[Bellemare et~al.(2016)Bellemare, Srinivasan, Ostrovski, Schaul,
  Saxton, and Munos]{bellemare2016unifying}
Marc Bellemare, Sriram Srinivasan, Georg Ostrovski, Tom Schaul, David Saxton,
  and Remi Munos.
\newblock Unifying count-based exploration and intrinsic motivation.
\newblock In \emph{Advances in Neural Information Processing Systems}, pp.\
  1471--1479, 2016.

\bibitem[Bellemare et~al.(2013)Bellemare, Naddaf, Veness, and
  Bowling]{bellemare2013arcade}
Marc~G Bellemare, Yavar Naddaf, Joel Veness, and Michael Bowling.
\newblock The arcade learning environment: An evaluation platform for general
  agents.
\newblock \emph{Journal of Artificial Intelligence Research}, 47:\penalty0
  253--279, 2013.

\bibitem[Chentanez et~al.(2005)Chentanez, Barto, and
  Singh]{chentanez2005intrinsically}
Nuttapong Chentanez, Andrew~G Barto, and Satinder~P Singh.
\newblock Intrinsically motivated reinforcement learning.
\newblock In \emph{Advances in neural information processing systems}, pp.\
  1281--1288, 2005.

\bibitem[Dhariwal et~al.(2017)Dhariwal, Hesse, Klimov, Nichol, Plappert,
  Radford, Schulman, Sidor, Wu, and Zhokhov]{dhariwal2017openai}
Prafulla Dhariwal, Christopher Hesse, Oleg Klimov, Alex Nichol, Matthias
  Plappert, Alec Radford, John Schulman, Szymon Sidor, Yuhuai Wu, and Peter
  Zhokhov.
\newblock Openai baselines.
\newblock \emph{GitHub, GitHub repository}, 2017.

\bibitem[Fran{\c{c}}ois-Lavet et~al.(2015)Fran{\c{c}}ois-Lavet, Fonteneau, and
  Ernst]{franccois2015discount}
Vincent Fran{\c{c}}ois-Lavet, Raphael Fonteneau, and Damien Ernst.
\newblock How to discount deep reinforcement learning: Towards new dynamic
  strategies.
\newblock \emph{arXiv preprint arXiv:1512.02011}, 2015.

\bibitem[Icarte et~al.(2018)Icarte, Klassen, Valenzano, and
  McIlraith]{icarte2018using}
Rodrigo~Toro Icarte, Toryn Klassen, Richard Valenzano, and Sheila McIlraith.
\newblock Using reward machines for high-level task specification and
  decomposition in reinforcement learning.
\newblock In \emph{International Conference on Machine Learning}, pp.\
  2112--2121, 2018.

\bibitem[Jaderberg et~al.(2016)Jaderberg, Mnih, Czarnecki, Schaul, Leibo,
  Silver, and Kavukcuoglu]{jaderberg2016reinforcement}
Max Jaderberg, Volodymyr Mnih, Wojciech~Marian Czarnecki, Tom Schaul, Joel~Z
  Leibo, David Silver, and Koray Kavukcuoglu.
\newblock Reinforcement learning with unsupervised auxiliary tasks.
\newblock \emph{arXiv preprint arXiv:1611.05397}, 2016.

\bibitem[Kingma \& Ba(2014)Kingma and Ba]{kingma2014adam}
Diederik~P Kingma and Jimmy Ba.
\newblock Adam: A method for stochastic optimization.
\newblock \emph{arXiv preprint arXiv:1412.6980}, 2014.

\bibitem[Lattimore \& Hutter(2011)Lattimore and Hutter]{lattimore2011time}
Tor Lattimore and Marcus Hutter.
\newblock Time consistent discounting.
\newblock In \emph{International Conference on Algorithmic Learning Theory},
  pp.\  383--397. Springer, 2011.

\bibitem[Lillicrap et~al.(2015)Lillicrap, Hunt, Pritzel, Heess, Erez, Tassa,
  Silver, and Wierstra]{lillicrap2015continuous}
Timothy~P Lillicrap, Jonathan~J Hunt, Alexander Pritzel, Nicolas Heess, Tom
  Erez, Yuval Tassa, David Silver, and Daan Wierstra.
\newblock Continuous control with deep reinforcement learning.
\newblock \emph{arXiv preprint arXiv:1509.02971}, 2015.

\bibitem[Machado et~al.(2017)Machado, Bellemare, Talvitie, Veness, Hausknecht,
  and Bowling]{machado17arcade}
Marlos~C. Machado, Marc~G. Bellemare, Erik Talvitie, Joel Veness, Matthew~J.
  Hausknecht, and Michael Bowling.
\newblock Revisiting the arcade learning environment: Evaluation protocols and
  open problems for general agents.
\newblock \emph{CoRR}, abs/1709.06009, 2017.

\bibitem[Martin et~al.(2017)Martin, Sasikumar, Everitt, and
  Hutter]{martin2017count}
Jarryd Martin, Suraj~Narayanan Sasikumar, Tom Everitt, and Marcus Hutter.
\newblock Count-based exploration in feature space for reinforcement learning.
\newblock In \emph{Proceedings of the International Joint Conference on
  Artificial Intelligence (IJCAI)}, 2017.

\bibitem[MattChanTK(2016)]{mattchantk_2016}
MattChanTK.
\newblock Mattchantk/gym-maze, 2016.
\newblock URL \url{https://github.com/MattChanTK/gym-maze}.

\bibitem[Mirowski et~al.(2016)Mirowski, Pascanu, Viola, Soyer, Ballard, Banino,
  Denil, Goroshin, Sifre, Kavukcuoglu, et~al.]{mirowski2016learning}
Piotr Mirowski, Razvan Pascanu, Fabio Viola, Hubert Soyer, Andrew~J Ballard,
  Andrea Banino, Misha Denil, Ross Goroshin, Laurent Sifre, Koray Kavukcuoglu,
  et~al.
\newblock Learning to navigate in complex environments.
\newblock \emph{arXiv preprint arXiv:1611.03673}, 2016.

\bibitem[Mnih et~al.(2013)Mnih, Kavukcuoglu, Silver, Graves, Antonoglou,
  Wierstra, and Riedmiller]{mnih2013playing}
Volodymyr Mnih, Koray Kavukcuoglu, David Silver, Alex Graves, Ioannis
  Antonoglou, Daan Wierstra, and Martin Riedmiller.
\newblock Playing atari with deep reinforcement learning.
\newblock \emph{arXiv preprint arXiv:1312.5602}, 2013.

\bibitem[Mnih et~al.(2016)Mnih, Badia, Mirza, Graves, Lillicrap, Harley,
  Silver, and Kavukcuoglu]{mnih2016asynchronous}
Volodymyr Mnih, Adria~Puigdomenech Badia, Mehdi Mirza, Alex Graves, Timothy
  Lillicrap, Tim Harley, David Silver, and Koray Kavukcuoglu.
\newblock Asynchronous methods for deep reinforcement learning.
\newblock In \emph{International Conference on Machine Learning}, pp.\
  1928--1937, 2016.

\bibitem[Ng et~al.(1999)Ng, Harada, and Russell]{ng1999policy}
Andrew~Y Ng, Daishi Harada, and Stuart Russell.
\newblock Policy invariance under reward transformations: Theory and
  application to reward shaping.
\newblock In \emph{ICML}, volume~99, pp.\  278--287, 1999.

\bibitem[Oh et~al.(2020)Oh, Hessel, Czarnecki, Xu, van Hasselt, Singh, and
  Silver]{oh2020discovering}
Junhyuk Oh, Matteo Hessel, Wojciech~M Czarnecki, Zhongwen Xu, Hado van Hasselt,
  Satinder Singh, and David Silver.
\newblock Discovering reinforcement learning algorithms.
\newblock \emph{arXiv preprint arXiv:2007.08794}, 2020.

\bibitem[OpenAI(2018)]{OpenAI_dota}
OpenAI.
\newblock Openai five.
\newblock \url{https://blog.openai.com/openai-five/}, 2018.

\bibitem[Oudeyer \& Kaplan(2009)Oudeyer and Kaplan]{oudeyer2009intrinsic}
Pierre-Yves Oudeyer and Frederic Kaplan.
\newblock What is intrinsic motivation? a typology of computational approaches.
\newblock \emph{Frontiers in neurorobotics}, 1:\penalty0 6, 2009.

\bibitem[Pitis(2019)]{pitis2019rethinking}
Silviu Pitis.
\newblock Rethinking the discount factor in reinforcement learning: A decision
  theoretic approach.
\newblock \emph{arXiv preprint arXiv:1902.02893}, 2019.

\bibitem[Reinke et~al.(2017)Reinke, Uchibe, and Doya]{reinke2017average}
Chris Reinke, Eiji Uchibe, and Kenji Doya.
\newblock Average reward optimization with multiple discounting reinforcement
  learners.
\newblock In \emph{International Conference on Neural Information Processing},
  pp.\  789--800. Springer, 2017.

\bibitem[Romoff et~al.(2019)Romoff, Henderson, Touati, Ollivier, Brunskill, and
  Pineau]{romoff2019separating}
Joshua Romoff, Peter Henderson, Ahmed Touati, Yann Ollivier, Emma Brunskill,
  and Joelle Pineau.
\newblock Separating value functions across time-scales.
\newblock \emph{arXiv preprint arXiv:1902.01883}, 2019.

\bibitem[Schmidhuber(2010)]{schmidhuber2010formal}
J{\"u}rgen Schmidhuber.
\newblock Formal theory of creativity, fun, and intrinsic motivation
  (1990--2010).
\newblock \emph{IEEE Transactions on Autonomous Mental Development}, 2\penalty0
  (3):\penalty0 230--247, 2010.

\bibitem[Schulman et~al.(2017)Schulman, Wolski, Dhariwal, Radford, and
  Klimov]{schulman2017proximal}
John Schulman, Filip Wolski, Prafulla Dhariwal, Alec Radford, and Oleg Klimov.
\newblock Proximal policy optimization algorithms.
\newblock \emph{arXiv preprint arXiv:1707.06347}, 2017.

\bibitem[Sherstan et~al.(2018)Sherstan, MacGlashan, and
  Pilarski]{sherstan2018generalizing}
Craig Sherstan, James MacGlashan, and Patrick~M Pilarski.
\newblock Generalizing value estimation over timescale.
\newblock \emph{Network}, 2:\penalty0 3, 2018.

\bibitem[Singh(1992)]{singh1992scaling}
Satinder~P Singh.
\newblock Scaling reinforcement learning algorithms by learning variable
  temporal resolution models.
\newblock In \emph{Machine Learning Proceedings 1992}, pp.\  406--415.
  Elsevier, 1992.

\bibitem[Stadie et~al.(2016)Stadie, Levine, and
  Abbeel]{stadie2015incentivizing}
Bradly~C Stadie, Sergey Levine, and Pieter Abbeel.
\newblock Incentivizing exploration in reinforcement learning with deep
  predictive models.
\newblock In \emph{ICLR}, 2016.

\bibitem[Sutton et~al.(1998)Sutton, Barto, et~al.]{sutton1998introduction}
Richard~S Sutton, Andrew~G Barto, et~al.
\newblock \emph{Introduction to reinforcement learning}, volume 135.
\newblock MIT press Cambridge, 1998.

\bibitem[Sutton et~al.(2000)Sutton, McAllester, Singh, and
  Mansour]{sutton2000policy}
Richard~S Sutton, David~A McAllester, Satinder~P Singh, and Yishay Mansour.
\newblock Policy gradient methods for reinforcement learning with function
  approximation.
\newblock In \emph{Advances in neural information processing systems}, pp.\
  1057--1063, 2000.

\bibitem[Sutton et~al.(2011)Sutton, Modayil, Delp, Degris, Pilarski, White, and
  Precup]{sutton2011horde}
Richard~S Sutton, Joseph Modayil, Michael Delp, Thomas Degris, Patrick~M
  Pilarski, Adam White, and Doina Precup.
\newblock Horde: A scalable real-time architecture for learning knowledge from
  unsupervised sensorimotor interaction.
\newblock In \emph{The 10th International Conference on Autonomous Agents and
  Multiagent Systems-Volume 2}, pp.\  761--768. International Foundation for
  Autonomous Agents and Multiagent Systems, 2011.

\bibitem[Tang et~al.(2017)Tang, Houthooft, Foote, Stooke, Chen, Duan, Schulman,
  DeTurck, and Abbeel]{tang2017exploration}
Haoran Tang, Rein Houthooft, Davis Foote, Adam Stooke, OpenAI~Xi Chen, Yan
  Duan, John Schulman, Filip DeTurck, and Pieter Abbeel.
\newblock \# exploration: A study of count-based exploration for deep
  reinforcement learning.
\newblock In \emph{Advances in Neural Information Processing Systems}, pp.\
  2750--2759, 2017.

\bibitem[Tieleman \& Hinton(2012)Tieleman and Hinton]{tieleman2012lecture}
Tijmen Tieleman and Geoffrey Hinton.
\newblock Lecture 6.5-rmsprop: Divide the gradient by a running average of its
  recent magnitude.
\newblock \emph{COURSERA: Neural networks for machine learning}, 4\penalty0
  (2):\penalty0 26--31, 2012.

\bibitem[Vaswani et~al.(2017)Vaswani, Shazeer, Parmar, Uszkoreit, Jones, Gomez,
  Kaiser, and Polosukhin]{vaswani2017attention}
Ashish Vaswani, Noam Shazeer, Niki Parmar, Jakob Uszkoreit, Llion Jones,
  Aidan~N Gomez, {\L}ukasz Kaiser, and Illia Polosukhin.
\newblock Attention is all you need.
\newblock In \emph{Advances in neural information processing systems}, pp.\
  5998--6008, 2017.

\bibitem[Veeriah et~al.(2019)Veeriah, Hessel, Xu, Rajendran, Lewis, Oh, van
  Hasselt, Silver, and Singh]{veeriah2019discovery}
Vivek Veeriah, Matteo Hessel, Zhongwen Xu, Janarthanan Rajendran, Richard~L
  Lewis, Junhyuk Oh, Hado~P van Hasselt, David Silver, and Satinder Singh.
\newblock Discovery of useful questions as auxiliary tasks.
\newblock In \emph{Advances in Neural Information Processing Systems}, pp.\
  9310--9321, 2019.

\bibitem[Xu et~al.(2018)Xu, van Hasselt, and Silver]{xu2018meta}
Zhongwen Xu, Hado~P van Hasselt, and David Silver.
\newblock Meta-gradient reinforcement learning.
\newblock In \emph{Advances in Neural Information Processing Systems}, pp.\
  2396--2407, 2018.

\bibitem[Xu et~al.(2020)Xu, van Hasselt, Hessel, Oh, Singh, and
  Silver]{xu2020meta}
Zhongwen Xu, Hado van Hasselt, Matteo Hessel, Junhyuk Oh, Satinder Singh, and
  David Silver.
\newblock Meta-gradient reinforcement learning with an objective discovered
  online.
\newblock \emph{arXiv preprint arXiv:2007.08433}, 2020.

\bibitem[Zahavy et~al.(2020)Zahavy, Xu, Veeriah, Hessel, Oh, van Hasselt,
  Silver, and Singh]{zahavy2020self}
Tom Zahavy, Zhongwen Xu, Vivek Veeriah, Matteo Hessel, Junhyuk Oh, Hado van
  Hasselt, David Silver, and Satinder Singh.
\newblock Self-tuning deep reinforcement learning.
\newblock \emph{arXiv preprint arXiv:2002.12928}, 2020.

\bibitem[Zheng et~al.(2018)Zheng, Oh, and Singh]{zheng2018learning}
Zeyu Zheng, Junhyuk Oh, and Satinder Singh.
\newblock On learning intrinsic rewards for policy gradient methods.
\newblock In \emph{Advances in Neural Information Processing Systems}, pp.\
  4644--4654, 2018.

\end{thebibliography}
\bibliographystyle{iclr2021_conference}


\end{document}